\documentclass[letterpaper]{article}

\usepackage{natbib,alifeconf}  
\usepackage{microtype}
\newcommand\algoname{DG-MAP-Elites}

\usepackage{hyperref}

\title{
Fast and stable MAP-Elites in noisy domains using deep grids}
\author{Manon Flageat$^{1}$ \and Antoine Cully$^{1}$ \\
\mbox{}\\
$^1$Imperial College London, Department of Computing, London, SW7 2AZ, UK. \\
Adaptive \& Intelligent Robotics Lab \\
a.cully@imperial.ac.uk} 

\begin{document}
\maketitle

\begin{abstract}
Quality-Diversity optimisation algorithms enable the evolution of collections of both high-performing and diverse solutions. These collections offer the possibility to quickly adapt and switch from one solution to another in case it is not working as expected. It therefore finds many applications in real-world domain problems such as robotic control. 
However, QD algorithms, like most optimisation algorithms, are very sensitive to uncertainty on the fitness function, but also on the behavioural descriptors. Yet, such uncertainties are frequent in real-world applications. 
Few works have explored this issue in the specific case of QD algorithms, and inspired by the literature in Evolutionary Computation, mainly focus on using sampling to approximate the "true" value of the performances of a solution. However, sampling approaches require a high number of evaluations, which in many applications such as robotics, can quickly become impractical.
\\ In this work, we propose Deep-Grid MAP-Elites, a variant of the MAP-Elites algorithm that uses an archive of similar previously encountered solutions to approximate the performance of a solution. We compare our approach to previously explored ones on three noisy tasks: a standard optimisation task, the control of a redundant arm and a simulated Hexapod robot. The experimental results show that this simple approach is significantly more resilient to noise on the behavioural descriptors, while achieving competitive performances in terms of fitness optimisation, and being more sample-efficient than other existing approaches.

\end{abstract}

\section{Introduction}
    \begin{figure}[ht]
    \centering
    \includegraphics[width=0.9\columnwidth]{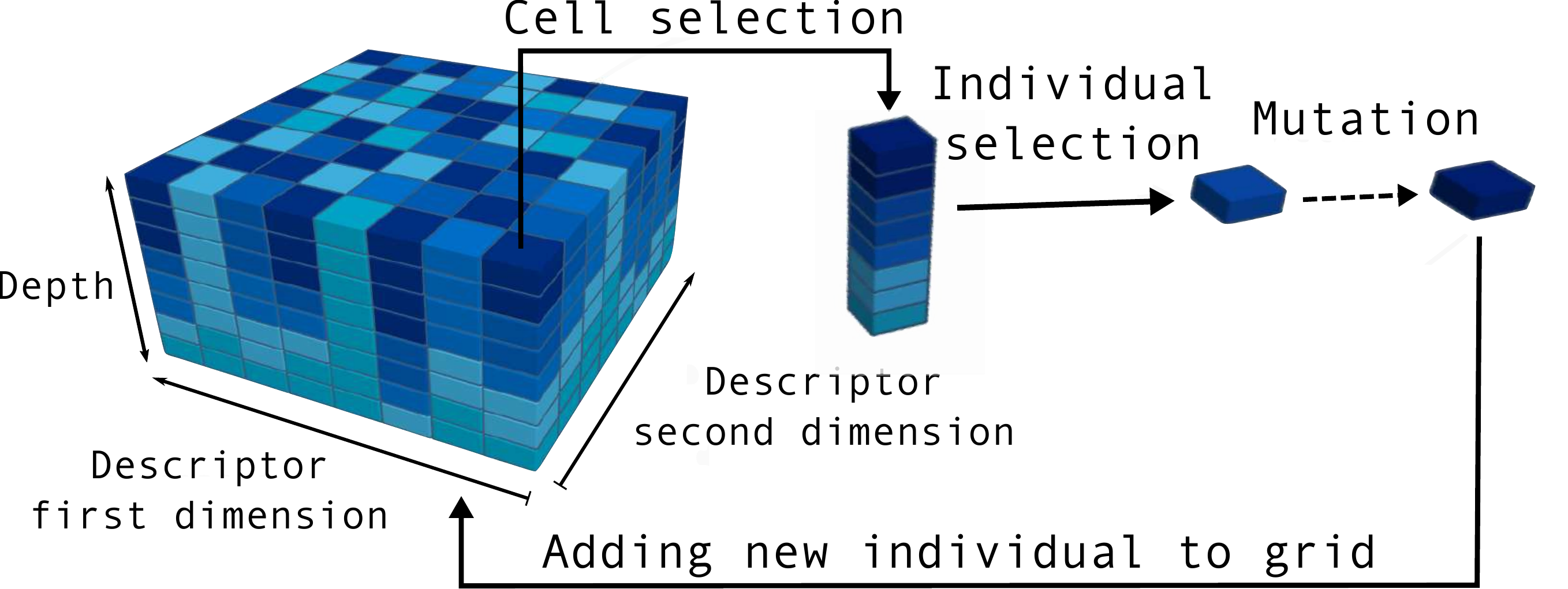}
    \caption{\algoname{} overview: at each generation, each parent individual is selected from the grid in two steps: first a cell is randomly selected, then an individual of this cell is selected with a probability proportional to its fitness; this parent individual is then mutated to generate an offspring, that will be added back to the cell it belongs to, randomly replacing one of the individuals contained in it.}
    \label{deep-grid}
\end{figure} 

Quality-Diversity (QD) algorithms \citep{pugh_quality_2016,cully_quality_2017} are a recently introduced class of evolutionary algorithms that aim at evolving repertoires of both diverse and high-performing solutions. These repertoires provide simple ways to quickly adapt to new or unseen situations by switching from one solution to another \citep{cully_robots_2015}. The diversity of these solutions is crucial to this adaptability.
To maintain this diversity, QD algorithm completes the fitness objective defining the problem, with the notion of novelty. This novelty is defined in a space characterising the practical effects of the solutions, known as the Behavior Descriptor (BD) space. 
QD algorithms have shown promising results in multiple domains. For example, they have been successfully applied to evolve diverse repertoires of gaits for robotic control \citep{cully_robots_2015, cully_evolving_2015, duarte2018evolution}; or in computer-aided design to generate multiple shapes of high-speed bicycles with various curvatures and volumes \citep{gaier_data-efficient_2018}; and used to generate procedural contents in video games \citep{gravina_procedural_2019}.

A  well-known  algorithm  of  this  family  is  MAP-Elites, introduced by \cite{mouret_illuminating_2015}. It uses a discretisation of the BD space into cells to maximise the diversity of the solutions.
MAP-Elites is elitist by design, which makes it highly sensitive to noises on the fitness function and the behavioural descriptor. If such noise causes a sub-optimal solution to get a high-fitness or to be considered particularly novel, it will be conserved in the repertoires and none of the solutions subsequently generated may manage to replace it. This would lead to keeping a sub-optimal solution in the final repertoire.

Several works focus on dealing with uncertainty on the fitness in the case of evolutionary algorithms (EA) \citep{yaochu_jin_evolutionary_2005, rakshit_noisy_2017}. However, few of them are directly applicable in the case of QD algorithms, and the existing works that used QD algorithms in uncertain environments mainly rely on sampling (repeated evaluations) to strengthen the performance approximation before adding solutions to the grid \citep{cully_hierarchical_2018}.

However, this often comes at the cost of a large number of evaluations or the application of more complex noise handling strategies \citep{justesen_map-elites_2019}.
However, sampling approaches require a high number of evaluations, and re-sampling a solution N times only divide the standard deviation of its approximate value by $\sqrt{N}$, \citep{rakshit_noisy_2017}. In addition, in domains such as robotics, evaluations of controllers are highly costly and time-consuming, and multiplying the number of evaluations can be really impractical \citep{chatzilygeroudis_survey_2020}.
It is only recently that a work investigated the use of adaptive sampling \citep{cantu-paz_adaptive_2004} with MAP-Elites algorithm to reduce the number of needed samples \citep{justesen_map-elites_2019}.

In this work, we introduce, Deep-Grid MAP-Elites (\algoname{}), a novel MAP-Elites variant which uses sub-populations to approximate the performances of cells. In \algoname{}, each cell of the traditional grid of MAP-Elites is extended to have a predefined depth to host several of the previously-encountered solutions. These sub-populations provide a way to implicitly sample an elite and have an approximation of its performance without any explicit sampling.
We compare this approach to the sample-based approaches detailed before on an optimisation task, and two robotic control tasks. We show that \algoname{} better approximates the value of the BD of the individuals, while reaching competitive quality performances. Moreover, our simple approach proves to be more data-efficient than sampling-based approaches.

    \begin{table*}[ht]
\small
\center{
\begin{tabular}{  c | c | c | c  } 
 Name &  Algorithm & Cell depth &  In-cell selection \\ 
 \hline \hline
  Baseline noise free & MAP-Elites without noise on the task & 1 & None \\
 \hline
  1 smpl Naive & MAP-Elites or 1-sampling explicit-averaging & 1 & None \\
 \hline
  50 smpl Naive & 50-samples explicit-averaging & 1 & None \\
 \hline
  Adapt & Adaptive-sampling & 1 & None \\
 \hline
  Adapt BD 10 & Adaptive sampling with drifting elites & 10 & First only \\
 \hline
  Deep grid 50 & \algoname{} & 50 & Fitness-prop. \\
\end{tabular}
\caption{Summary of the algorithms variants compared in the experiments.}
 \label{algo_table}
}
\end{table*} 

\section{Background}
    \subsection{Quality-Diversity and MAP-Elites algorithms}
QD algorithms aim at evolving containers of solutions that are both diverse and high-performing \citep{pugh_quality_2016}.
They are based on the definition of Behaviour Descriptor (BD) space, which defines a second characterisation of a solution, in addition to its fitness. The BD space is often defined using dimensions relevant to the goal of the optimisation process, and the BD of a solution is a projection of its effect on these dimensions. 
For example, in robotics, this space can be the final position of the robot reached thanks to the solution \citep{cully_quality_2017}, or alternatively the use of joints it induces \citep{cully_robots_2015}.
The BD space can either be hand-designed \citep{pugh_quality_2016} or automatically determined \citep{cully2019autonomous, nguyen2016understanding}. 
For a given solution, the genotype is mapped to a phenotype, which is then evaluated, and its actions and their consequences are mapped to a BD and quantified in term of performance with a fitness value. 
Solutions can then be compared according to their fitness or to their novelty, namely their ability to reach a less explored part of the BD space. This is used to build the container of solutions.

There are currently two leading QD approaches, which share multiple aspects, and can be integrated in a general framework as proposed in \cite{cully_quality_2017}. 
The first one is Novelty Search with Local Competition (NSLC), proposed in \cite{lehman_evolving_2011}. Its core idea is to search for solutions that are both novel and locally high-performing. This is done with a multi-objective optimisation algorithm and by storing the encountered solutions according to their BD in a "novelty archive", which is used to compute the novelty score and local performance score. The second one, MAP-Elites, which is the focus of this work, has been proposed in \cite{mouret_illuminating_2015}. It is based on the discretisation of the BD space into a grid, in which each cell can contain one individual, called an elite. MAP-Elites aims at finding the best possible solution for each of these cells. One iteration of MAP-Elites can be summarised as: 1. randomly selecting a given number of individuals from the grid, 2. applying mutation to generate offspring, and evaluating them, 3. adding offspring that either populate an empty cell or outperform an existing elite. 
This simplicity of implementation makes MAP-Elites a good candidate for application of Quality-Diversity optimisation.

    \subsection{Related works in EA} 

\paragraph{Optimisation of noisy fitness functions}
A fitness function is defined as noisy or uncertain, when only unreliable measurements of its value can be acquired, meaning that multiple evaluations of the same solution would lead to different fitness values. The performance of a solution $\vec{x}$ given by a noisy fitness function can be formulated as: $f_{noisy}(\vec{x}) = f(\vec{x}) + \epsilon(\vec{x})$, with $f(\vec{x})$ the effective value of the fitness of $\vec{x}$, and $\epsilon(\vec{x})$ sampled from a distribution that may be dependent on $\vec{x}$. 

The optimisation of noisy fitness functions is a common concern in evolutionary algorithm literature and multiple works have attempted to propose solutions to this problem, a survey of these works can be found in \cite{yaochu_jin_evolutionary_2005} and \cite{ rakshit_noisy_2017}.
The most common classification of the proposed approaches contrasts Explicit-averaging and Implicit-averaging approaches. In Explicit-averaging, each individual is re-sampled multiple times and the mean performance value is used as the performance of the individual. In Implicit-averaging, the population size is increased, raising the probability that similar individuals are encountered in the population and implicitly sample the solutions. 

\noindent \textbf{The explicit-averaging} approach is the most straightforward approach to noisy optimisation; it samples each solution a fixed-number of times to get an average value of its fitness. However, this approach may waste samples on non-promising solutions, thus multiple works have focused on developing "Adaptive-sampling" approaches that aim at economising samples by distributing them more wisely among the individuals. For example, \cite{aizawa_sequential_1994} propose to sample more highly-performing individuals, or to distribute samples based on variance values. In another work, as \cite{cantu-paz_adaptive_2004}, focused on reducing the number of samples to the minimum required to discriminate between individuals. 

\noindent \textbf{The implicit-averaging} approach is rather based on population than on sampling. In its raw form, it consists in increasing the population size, to implicitly sample the fitness landscape with a higher number of close similar individuals. It has proven to be efficient in theory \citep{miller1996genetic}. Thus, a few works have focused on approaches inspired by this idea, such as \cite{branke_efficient_2001} which averages over the neighbours of an individual to evaluate its true fitness value. Similarly, the work by \cite{branke_creating_1998} proposes to maintain an archive of past neighbouring individuals to approximate the value of the current individuals. This approach was initially introduced for robust optimisation rather than noisy optimisation; however, similar tools have been developed to tackle both of these issues and they are often confused or voluntarily merged.

The two methods were alternatively proven to be better depending on the type of task and the structure of the noise \citep{beyer_toward_1993,back_evolution_1994-1}. Explicit-averaging has a high sample-cost, while Implicit-averaging induces a trade-off between noise-handling and selection-pressure. Explicit-sampling is often preferred due to this last limitation and the simplicity of interpretation of sample-based approaches. However, Implicit-averaging is sometimes chosen as it is more sample-efficient by construction.
    \subsection{Related works in QD} 

In QD, noisy domains present an additional challenge as both the fitness and the BD can be noisy. Multiple works use explicit-averaging approaches to handle uncertainty, as in \cite{cully_hierarchical_2018} and \cite{gomes_approach_2018}, where the BD is evaluated 100 times per individual. 
However, to our knowledge, only one work by \cite{justesen_map-elites_2019} proposes a more complex strategy to handle noise in QD algorithms. Their approach is inspired by the Adaptive-sampling approach and applied to MAP-Elites. Its core principle is that an individual can replace an elite in its cell only if it is still better after having been sampled the same number of times as the elite. An important idea is that each time a new individual proves less performing than an elite, this elite is re-sampled. 
One difficulty with this method is that some elites may prove to belong to another cell than the one they are in and need to be moved, leading to "drifting elites". Thus, two versions of the Adaptive-sampling algorithm are proposed: the first one does not take into account these drifting elites, and keeps the elites where they are initially evaluated; and the second one allows an elite to drift to its actual cell and to be compared to a potentially already present elite. However, in this second approach, the drift of elites leads to empty cells that represent a loss of performance for the optimisation process. To reduce this effect, the second algorithm grid is improved with a "depth" for each cell, that keeps the following best individuals. Therefore, when an elite drift, the second-best individual can take its place.
We reproduce the results of this work and use these two algorithms, as well as the Explicit-averaging approach, as references to compare the performances of our approach.
    \begin{figure*}[ht]
\begin{center}
\includegraphics[width=0.88\textwidth]{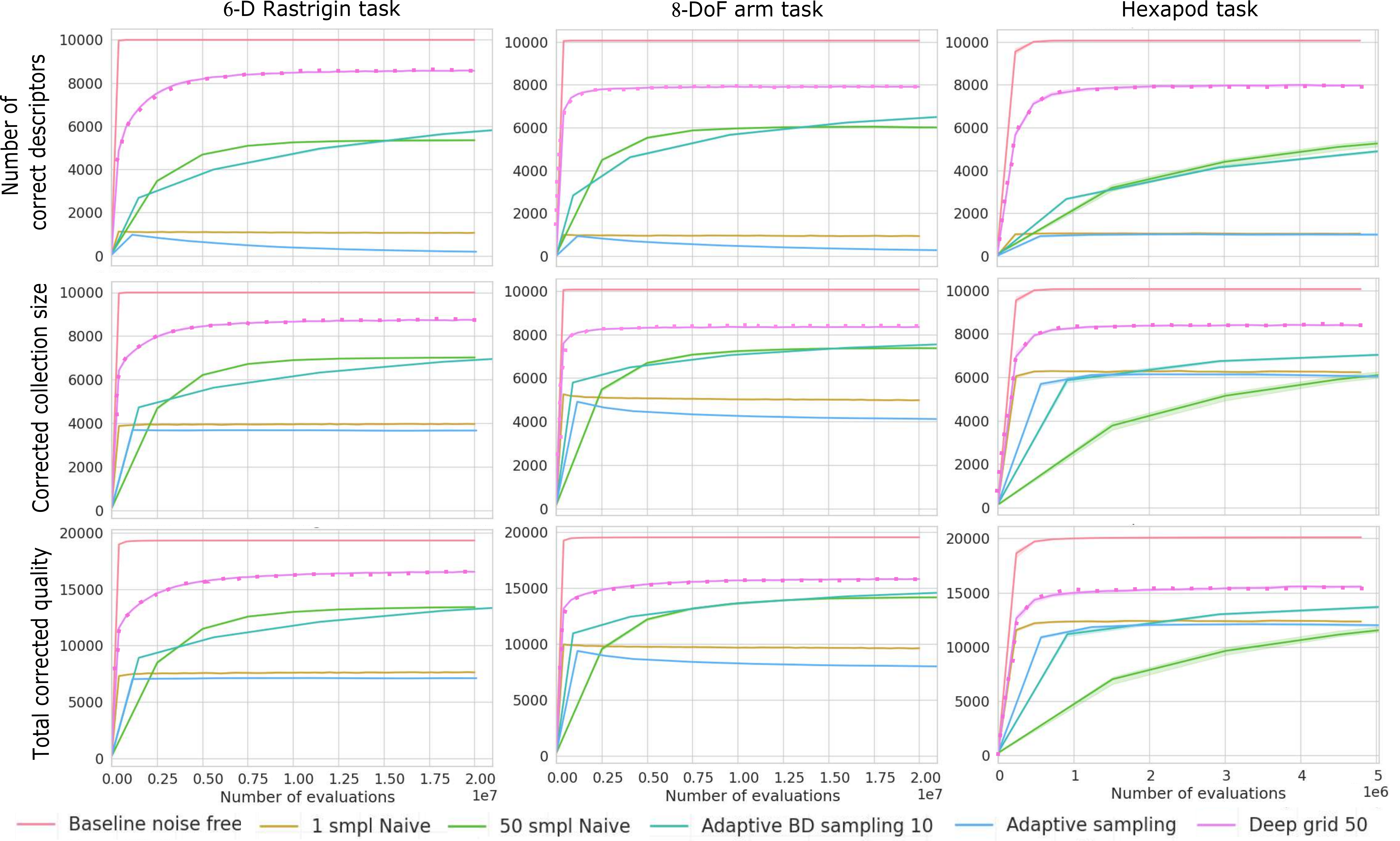}
\caption{Comparison of the algorithms on the three tasks according to the three metrics: number of correct descriptors (top), corrected collection size (middle) and total corrected quality (bottom). 
The resolution differences between the algorithms comes from that fact the data are saved periodically after a certain number of generations. Algorithms doing numerous evaluations during this period, like 50 smpl Naive, have a lower resolution than algorithms doing fewer evaluations, like Deep grid 50. 
}
\label{graphs}
\end{center}
\end{figure*} 
    
\section{\algoname{} algorithm}
    \algoname{} is an extension of MAP-Elites for noisy domain optimisation. The core idea of \algoname{} is to use the individuals encountered during the optimisation process to approximate the values of the fitness and the BD of the best individuals.
This approach shares similarities with the works based on neighbouring-archive proposed for EA in \cite{branke_creating_1998}. 
In the case of QD algorithms, though, it is more relevant to define the notion of neighbourhood in the BD-space instead of the phenotype or genotype space. Indeed, in QD algorithms,  the distance between solutions is often defined in the BD-space. 
For instance, MAP-Elites specifies a neighbourhood in the BD-space for each elite by defining the cells of the grid. \algoname{} uses this segmentation of the BD-space and sets up one local-archive of past-individuals in each cell. Using local-archives allows to quickly identify past-individuals that are of interest to approximate a given elite.
These archives use a maintenance-scheme to conserve relevant past-solutions coherent with the current population, similarly to the improvement that has been investigated for the \cite{branke_creating_1998} neighbouring-archive work by \cite{kruisselbrink_archive_2010}.

\algoname{} is based on the assumption that the fitness landscape of elites with respect to BD is locally smooth. This assumption has already been done in previous QD work such as \cite{cully_robots_2015}. In the case of \algoname{}, it allows the use of an individual's neighbourhood to approximate its fitness value.

The main steps of MAP-Elites algorithm are kept in \algoname{}: a solution is selected from the grid, mutated and its offspring is evaluated before being added back to the grid. The selection operator, designated in \cite{cully_quality_2017} as selector, and the grid management scheme specific to \algoname{} are introduced below.
The main difference between \algoname{} and MAP-Elites algorithm is the addition of a "depth": each cell of \algoname{} grid contains $D$ individuals, instead of a unique elite, $D$ being the depth of the cell. These deep-cells contain individuals that have been encountered during the optimisation process with a BD that belongs to the cell, genuinely or due to the noise.

As the BD-space is a projection, multiple solutions with distinct fitness values may belong to the same cell. Therefore, one key aspect of \algoname{} is that, over a large number of generations, each cell will slowly be populated by potential replications of similar solutions. 
These sub-populations implicitly sample one solution belonging to the cell. Then, the goal is to ensure that it is the optimal solution, by isolating high-performing solutions, while  avoiding to collapse too early to replications of sub-optimal solutions. This aspect motivated the choice of the selection operator and the container management scheme described below.

A graphical representation of \algoname{} can be found in Fig.~\ref{deep-grid}.

\subsection{\algoname{} selector} 
One key aspect of \algoname{} is to keep MAP-Elites elitism, while avoiding that luckily-high-performing individuals populate the cell. Therefore, the selector of \algoname{} has been chosen to promote highly-performing individuals while giving a chance to less successful individuals that may have been unlucky during the evaluation. 

It is defined in two consecutive steps. First, at a grid-scale, \algoname{} selector is similar to MAP-Elites: a cell is randomly chosen among the cells that contain at least one individual. This rule is crucial to QD approach as it enables to equitably explore all areas of space and to promote diversity. Second, inside the selected cell, an individual is selected among those contained in the cell based on a fitness-proportional selector. This in-cell rule avoids focussing on a unique elite that could have got its high-fitness score by chance, while keeping a selective pressure biased toward high performing individuals. The two steps of this selector are performed successively when selecting an individual: first, a cell is randomly chosen, second, an individual from this cell is selected using a fitness-proportional selector.

\subsection{\algoname{} container maintenance-scheme} 

The main difference between \algoname{} and MAP-Elites is the addition of a depth in each cell. The maintenance scheme of the container is thus really different from the one used in MAP-Elites. 
One key aspect of \algoname{} is to avoid that lucky or unstable individuals populate the cells. Therefore, the main motivation of the container maintenance-scheme is to artificially encourage individuals that can reproduce their performance. This can be done by systematically re-questioning individuals, thus, forcing them to be highly-reproducible to be kept in the cell. A straightforward implementation of this idea is to always add newly encountered individuals to the container, randomly replacing one individual already in the cell. In \algoname{}, as in MAP-Elites, offspring are generated from parents taken exclusively from the grid. 
Thus, any offspring is a near replication of an individual already in the grid. An individual correctly-evaluated in the right cell is more likely to yield to an offspring belonging to the same cell, than the same individual that has been wrongly-evaluated in an incorrect cell.
Moreover, if the correctly-evaluated individual is also high-performing, it will be selected more often, following the selector rule. As these stable and performing individuals get selected and mutated, they will slowly become more predominant in their cell and replace any other sub-optimal individual inside the cell. Therefore, randomly replacing individuals inside the cell allows \algoname{} to slowly eliminate uncertain and sub-performing individuals.

To summarise, in \algoname{}, all individuals encountered during the optimisation process are added to the container in the cell they belong to. As long as this cell is not full, they are simply added to it without discarding any existing solutions. However, as soon as the cell is full, the new solutions replace individuals randomly chosen among the ones already in the cell. The respective fitness of the individuals and the time-of-appearance are not taken into account to remove any potential bias.

\subsection{\algoname{} implementation} 
\algoname{} only adds a single hyper-parameter, which is the depth of the grid.
Furthermore, the container maintenance scheme does not depend on any fitness value and implementing a fitness proportional selector can be done without any need for an ordering, thus we implemented a really-simple container with disorganised individuals inside the cell.
In short, transforming a traditional MAP-Elites implementation into a \algoname{} only requires 3 simple changes: 1) increasing the number of dimensions of the MAP-Elites grid to add depth in the grid, 2) adding the second step of the selection operator: after randomly selecting a cell, using a fitness proportionate selection to select one individual, and 3) changing the addition condition into the replacement of a randomly selected individual in the cell. 
The complete \algoname{} algorithm is summarised in Algorithm \ref{algorithm}.

\begin{algorithm}
\small
 \textbf{Input: }$N_{gen}$ generations, $N$ population size\;
 Initialisation of the grid\;
 \For{$generation=1,2,\ldots, N_{gen}$}
 {
   // \textit{Select the parents from the grid} \\
   \For{parent $p=1,2,\ldots, N$}
   {
     $cell_p$ =  random selection of cell in grid \;
     $indiv_p$ = fitness-prop selection of indiv in $cell_p$ \;
     $pop_{parents}(p)$ = $indiv_p$\;
   }
   // \textit{Get and evaluate offspring} \\
   $pop_{offspring}$ = mutate($pop_{parents}$)\;
   $BD$, $fit$ = evaluate($pop_{offspring}$)\;
   // \textit{Add offspring to the grid} \\
   \For{offspring $o$ in $pop_{offspring}$}
   {
      $cell_o$ = find cell for $BD(o)$ \;
      \eIf{$cell_o$ is empty or not full}{
        add $o$ to the cell \;
        }{
        $indiv$ = random selection in $cell_o$\;
        replace $indiv$ with $o$ \;
        }
   }
 }
 \textbf{return} grid
 \caption{\algoname{} algorithm}
 \label{algorithm}
\end{algorithm} 

\section{Experimental setup}
    We compare the performance of \algoname{} to four other approaches on three tasks commonly studied in QD works: one standard optimisation task, one control task on a robotic arm and one more complex control task on a hexapod robot. Similarly to what has been done in \cite{justesen_map-elites_2019}, we add Gaussian random noise on both the fitness: $\mathcal{N}(0,\,0.05)$, and the descriptor: $\mathcal{N}(0,\,0.01)$ of the solutions in each of the three tasks.

    \subsection{Compared algorithms and baselines}

On those three tasks, we compare three types of algorithms: Explicit-averaging algorithms (designated as "Naive" in the results), Adaptive-sampling algorithms implemented following the approach proposed in \cite{justesen_map-elites_2019} (designated as "Adapt"), and \algoname{} algorithm (designated as "Deep").

\noindent \textbf{Explicit-averaging algorithms} "naively" sample offspring individuals N times before comparing them to the existing individuals in the container, N being fixed as a hyper-parameter. We compare the same algorithm with $N=1$, which correspond to the traditional MAP-Elites algorithm evaluating each individual once; and $N=50$, which sample a high-number of times each offspring, being therefore highly costly in evaluations. These two approaches act as two extreme baselines: the former showing no noise-handling strategy and the latter a robust but particularly data-expensive noise-handling strategy.

\noindent \textbf{Adaptive-sampling algorithms} balance this cost by distributing the samples on the most promising solutions. In this work, we implemented the two adaptive-sampling algorithms proposed in \cite{justesen_map-elites_2019}: Adaptive-sampling ("Adapt") which increases the number of evaluation of each individual with time to refine the approximation as the algorithm converges; and Adaptive-sampling with drifting elites ("Adapt BD"), which is using the exact same principle but allows the best individual of a cell to drift to another cell if its descriptor appears to have been misjudged. This second algorithm also keeps multiple solutions per cell, like \algoname{}. However, this is used to quickly replace drifting individuals and therefore is not considered in the selector. To make a fair comparison, we used in preliminary experiments Adaptive-sampling with depth $d=50$ to match the depth chosen for \algoname{} and the number of samples used in the Explicit-sampling algorithm. However, the performance of Adaptive-sampling has proven to be better with the value of $d=10$ used in the original work by \cite{justesen_map-elites_2019}.

\noindent \textbf{\algoname{} algorithm} is the approach proposed in this work. Here we compare depth $d=50$ to match the same order of magnitude as naive sampling.

These algorithms and their parameters are summarised in Table~\ref{algo_table}.
We also add a noise-free baseline which gives an absolute reference for each analysis in the absence of uncertainty. This baseline consists of a traditional MAP-Elites algorithm sharing the same parameters as the other variants but optimising the task without any noise.
    \begin{figure*}[ht]
\begin{center}
\includegraphics[width=0.97\textwidth]{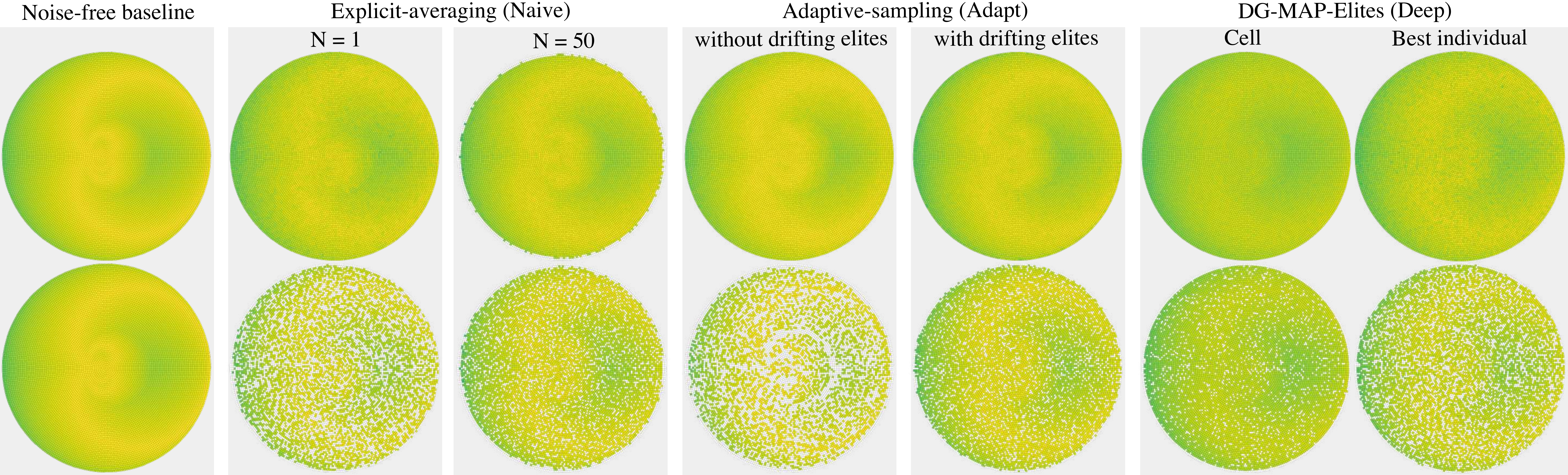}
\caption{
Container plot for the arm-control task of all algorithms.
The behaviour descriptor space, discretised in cells, corresponds to the position of the gripper in two dimensions. The fitness is computed from the variance of the angles in the joints and given by the colour: the brighter the better.
In the top container graphs, the fitness of each cell is re-evaluated $N_{repeat}=50$ times. In the bottom container graphs, the fitness and the BD are re-evaluated $N_{repeat}=50$ times, and the cells are placed where they belong to, corresponding to the corrected-containers.
As \algoname{} is the only population-based approach, the quality of its cell is pulled down by averaging, so we add the best-individual container for comparison purpose. 
}
\label{arm_archives}
\end{center}
\end{figure*} 
    \subsection{6-D Rastrigin domain} 

The first task is a 6-D Rastrigin domain subject to Gaussian noise on the fitness and the BD.
\\ \textbf{Genotype: } a solution is given by the 6 real-valued variables used to compute the fitness function: $\vec{x} =(x_i)_{1<i<6}$.
\\ \textbf{Descriptor: } each solution is described by the values of the first 2 of the 6 variables: $(x_1, x_2)$.
\\ \textbf{Fitness: } the fitness is given by the 6-D Rastrigin function: 
$$f(\vec{x})=-60 - \sum_{i=1}^{6}{(x_i^2-10 \cos{(2 \pi x_i)})}$$

\subsection{Simulated 8-degrees-of-freedom arm control}

The second experiment is the control of a simulated 8-Degree-of-Freedom (8-DoF) robotic arm inspired by \cite{cully_quality_2017}. The aim of this task is to find how to access all the points reachable by the arm, while minimising the variance between the different angles applied in each of its DoF. Gaussian noise is also added on the fitness and BD.
\\ \textbf{Genotype: } a solution is defined by a set of real-valued angles, one for each of the 8-DoF of the arm: $\vec{\theta}=(\theta_i)_{1<i<8}$.
\\ \textbf{Descriptor: } a controller is described by the $(x,y)$ position of the end-effector of the arm, after applying the control values in the joints.
\\ \textbf{Fitness: } the performance of a solution is given by the negative variance of the angles $(\theta_i)_{1<i<8}$: 
$$f(\vec{\theta})=-V((\theta_i)_{1<i<8}) = -\frac{1}{8}\sum_{i=1}^{8}{(\theta_i-\overline{\theta})^2}$$

\subsection{Simulated hexapod robot control}

The third task, from \cite{cully_evolving_2015}, aims at learning how to walk in every direction following circular trajectories with a simulated hexapod robot with 3-DoF per leg.
\\ \textbf{Genotype: } a solution is a controller defined by a set of periodic functions, applied in each DOF of the robot and characterised by their amplitude, phase, and duty cycle. 
\\ \textbf{Descriptor: } a controller is described by the final $(x,y)$ position of the robot.
\\ \textbf{Fitness: } the fitness is the negative value of the angle difference between the final orientation of the robot $\alpha(\vec{x})$ and the tangent of the circle through its start and end position $\beta(\vec{x})$: 
\begin{center} $f(\vec{x})=-| \alpha(\vec{x}) - \beta(\vec{x}) | $ \end{center}
    \subsection{Metrics}

Given the intrinsic differences of each approach, a crucial point of this work was to design an appropriate way to compare them. The main difference to take into account is the presence of a depth, but also its usage, which differs between Adaptive-sampling with drifting elites and \algoname{}. 

Thus, we compare the algorithms at a cell-level instead of an individual-level. To define the effective fitness and the BD of a cell, we call $N_{repeat}$ times the in-cell selector as a way to sample individuals (the approaches without depth always return the only individual contain in each of their cells). This method enables us to consider in the same framework approaches without depth (Explicit-averaging, and Adaptive-sampling without drifting elites) as well as methods using the notion of depth (Adaptive-sampling with drifting elites, \algoname{}). It is important to note that Adaptive-sampling without drifting elites uses the depth to substitute elites when they drift. The in-cell selector always returns the best individual per cell. Conversely, the in-cell selector of \algoname{} returns different individuals based on their fitness when called multiple times. 

The values based on the $N_{repeat}$ evaluations are considered as the ground truth for this cell-entity; but might reveal that some of the cells are mis-evaluated and need to be moved to neighbouring cells. We thus use a \textbf{corrected container} in which each cell is moved to its ground-truth BD-value, and if multiple cell-entities belong to the same cell, the best one is preserved while the others are discarded (like in \cite{justesen_map-elites_2019}). This leads to corrected containers that may present gaps in place of some cells. This corrected container is used to compute all the following quantitative metrics.

We compare the six algorithms with three metrics, among which two are the same as the ones used by \cite{justesen_map-elites_2019}. The first of these metrics is the \textit{Corrected-collection size}, which corresponds to the total number of filled-cells in the corrected container. It aims at quantifying the diversity of the solutions found by the algorithm and is comparable to the collection size metric commonly used to analyse QD algorithms \citep{pugh_quality_2016}.
Similarly, the second metric is the \textit{Total normalised corrected quality}: the sum of the normalised quality of all cells in the corrected container, that quantifies the quality of the solutions and is comparable to the traditional QD-score \citep{pugh_quality_2016}.
These two metrics are used in \cite{justesen_map-elites_2019}. We add a third metric, the \textit{Number of correct BD} to quantify the stability of the solutions found by the algorithm. It corresponds to the number of cells which corrected-BD belongs to the exact same cell, and it therefore differs from the corrected-collection size as it does not take into account cells that may have drifted to a new position left empty.
Moreover, an important resource that all algorithms try to minimise is the number of evaluations, thus, we represent all our metrics with respect to it instead of the number of generations. We also run all the algorithms for a fixed number of evaluations instead of a fixed number of generations as it is usually the case for QD-algorithms.

\subsection{Implementation and hyperparameters}

For a fair comparison, all the algorithms use the same hyperparameter values. The only exceptions are the depth, which is specific to each algorithm; and the mutation-rate. As \algoname{} is based on the reproduction of similar solutions, it depends on the mutation rate value; we tune it to $5\%$, but we keep its value to $10\%$ for the other approaches following heuristics (no crossover used).
The MAP-Elites grid definition is shared across the algorithms but varies between the tasks: the Rastrigin task uses a Cartesian grid with $10000$ cells; and the two robotic tasks a Polar grid of $10062$ cells (similar to the one proposed in \cite{gaier2020automating}), as these two tasks imply a circular distribution of the individuals. The Rastrigin and arm tasks are run for $2*10^7$ evaluations and replicated $50$ times. The Hexapod task is run for $5*10^6$ evaluations and replicated $10$ times, due to the high computational cost of this experiment (one replication takes 50 hours on 32 CPUs).
The implementation of all tasks and algorithms is based on the Sferes2 library by \cite{Mouret2010}, and the hexapod  control task uses the Dart simulator by \cite{Lee2018}. The source code and a Singularity container with the corresponding environment for replication can be found at {\footnotesize \url{https://github.com/adaptive-intelligent-robotics/Deep-Grid\_MAP-Elites.git}}.
    \begin{figure*}[ht]
\begin{center}
\includegraphics[width=0.97\textwidth]{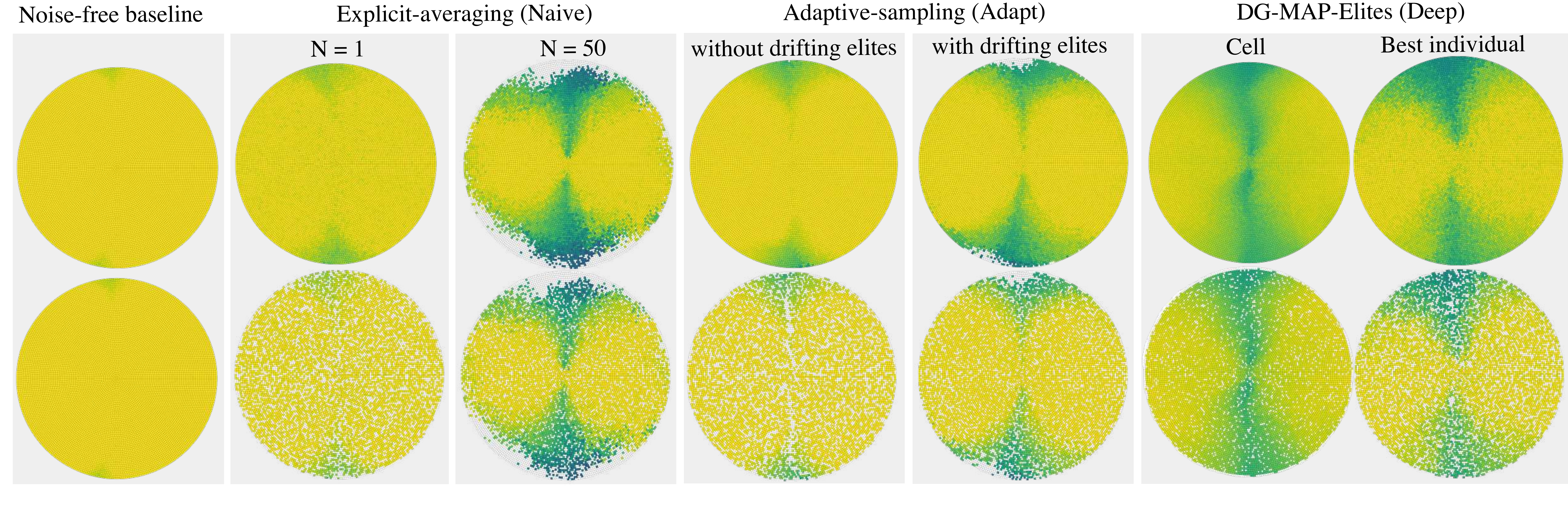}
\caption{Container plots of the Hexapod control task, similar to Figure \ref{arm_archives}. The behaviour descriptor space, discretised in cells, is the final position of the Hexapod robot along the two first dimensions; and the fitness, given by the colour, its final orientation.}
\label{hexa_archives}
\end{center}
\end{figure*} 
    
\section{Results}
    The comparison of the six algorithms on the three tasks is summarised in Fig~\ref{graphs}. Additionally, a graphical representation of the containers for the 8-DoF arm and Hexapod control tasks can be found in Fig~\ref{arm_archives} and \ref{hexa_archives} respectively. 
The container of each algorithm is first represented after having evaluated the fitness $N_{repeat}=50$ times. The same container is then represented again, after moving the solutions according to the corrected-BD.  
As explained above, \algoname{} is evaluated by averaging the sampled individuals from each cell. These in-cell sub-populations constitute its strength as it allows for more stability in the container. Therefore, the following analysis compares the cell-approach described in the previous section.
To offer a different perspective, we also show the same plots with the best individual of each cell. This additional plot illustrates that while \algoname{} has a slightly lower average performance it also finds high-performing individuals in each cell. 

Fig~\ref{graphs} show that, across tasks, the relative performances of the algorithms are similar. Furthermore, the analysis of these results with the Wilcoxon Ranksum test show that the performance differences between the algorithms are statistically significant (all p-values are $< 0.003$).

The quantitative analysis of Fig~\ref{graphs} highlights the limitations of traditional MAP-Elites, or 1-sample Explicit-averaging, in face of uncertainty. Its performance in the \textit{Number of correct descriptors} metric shows that less than $20\%$ of the elites are correctly placed on the grid. The \textit{Corrected collection size} metric and \textit{Total corrected quality} metrics result from this low stability of the elites. They show that it directly impacts the algorithm quality and diversity performances.

Interestingly, Adaptive-sampling shows even worse performance than traditional MAP-Elites. This is due to its impossibility to move elites from one cell to another when their BD is mis-evaluated at the first evaluation. When there is noise on the BD, as in this case, this approach "blocks" some of the cells with individuals that are truly high-performing but do not belong to the cell they are in. These individuals cannot be beaten by lucky individuals belonging to a closer cell, as the adaptive criterion counteracts luck on the fitness. This effect can be observed qualitatively on the container plots in Fig~\ref{arm_archives}: Adaptive-sampling without drifting elites has a thicker and brighter central arabesque than any other algorithms, that leaves a partially-empty area in the re-evaluated archive.

In comparison, Adaptive-sampling with drifting elites and Explicit-averaging with $N=50$ manages to get similar good-performance. These performances are reached faster by Adaptive-sampling thanks to its sample-management scheme. Overall, Adaptive-sampling with drifting elites shows better results than the Explicit-averaging approach each time the adaptive scheme allows it to sample elites more than $N=50$. 

On each metric and across all tasks, \algoname{} gets the highest score of the algorithms. It mis-evaluates only $20\%$ of the cells of the container and this high stability allows it to keep a high collection size. \algoname{} also gets higher performance on the \textit{Total corrected quality} metric, while the difference to other algorithms is less pronounced than for other metrics. This highlights the limitation of \algoname{} which quality performance is pulled down by averaging over a population (visible in Fig.~\ref{arm_archives} and \ref{hexa_archives}). However, its convergence speed proves to be particularly fast in terms of number of evaluations, especially when compared to Explicit-averaging and Adaptive-sampling approaches, which spend a high number of evaluations on each individual.

\section{Conclusion and discussion}
    In this work, we present \algoname{}, a sub-population-based variant of MAP-Elites algorithm for noisy domain optimisation. \algoname{} makes use of individuals previously encountered during the optimisation process to approximate the value of current individuals and avoid using re-evaluation. It notably expands its container rules to avoid favouring lucky individuals and allow for solutions that initially seem less efficient but prove unlucky, to emerge.
We show that \algoname{} is faster and allows for higher stability of the archive solutions while achieving competitive quality performance to the sample-based approaches. 

This work presents the performance of \algoname{} in the face of an invariant noise structure. Thus, an interesting future work would be to apply this approach to more complex noise structures. For example, solution-dependant noise would allow to study the behaviour of \algoname{} approach to handle situations in which the best-performing solutions may not be the less noisy ones, which would correspond to study \algoname{} approach for robust QD optimisation.
The simplicity of \algoname{} also opens a large number of research directions. For instance, improving its convergence speed and the overall quality of the solutions, by introducing more advanced container rules. However, avoiding undesired bias or influence coming from the noise remains a challenging task.

From a broader perspective, the range of applications of the approaches compared in this work is quite different: population-based approaches easily scale to tasks with high-cost of evaluation and a need for good mean-performance, while sampling-based approaches easily apply to quick-evaluation tasks, with a need for high one-shot performance.

\bibliographystyle{apalike}
\bibliography{biblio.bib} 

\begin{thebibliography}{}

\bibitem[Aizawa and Wah, 1994]{aizawa_sequential_1994}
Aizawa, A. and Wah, B. (1994).
\newblock A sequential sampling procedure for genetic algorithms.
\newblock {\em Computers \& Mathematics with Applications}, 27(9):77--82.

\bibitem[Beyer, 1993]{beyer_toward_1993}
Beyer, H.-G. (1993).
\newblock Toward a {Theory} of {Evolution} {Strategies}: {Some} {Asymptotical}
  {Results} from the (1,+ $\lambda$)-{Theory}.
\newblock {\em Evolutionary Computation}, 1(2):165--188.

\bibitem[Branke, 1998]{branke_creating_1998}
Branke, J. (1998).
\newblock Creating robust solutions by means of evolutionary algorithms.
\newblock In {\em Parallel {Problem} {Solving} from {Nature}}, pages 119--128.

\bibitem[Branke et~al., 2001]{branke_efficient_2001}
Branke, J., Schmidt, C., and Schmeck, H. (2001).
\newblock Efficient {Fitness} {Estimation} in {Noisy} {Environments}.
\newblock In {\em {Genetic} and {Evolutionary} {Computation} Conference}, pages
  243--250.

\bibitem[Bäck and Hammel, 1994]{back_evolution_1994-1}
Bäck, T. and Hammel, U. (1994).
\newblock Evolution {Strategies} {Applied} to {Perturbed} {Objective}
  {Functions}.
\newblock In {\em {IEEE} {World} {Congress} of {Computational} {Intelligence}},
  pages 40--45.

\bibitem[Cantu-Paz, 2004]{cantu-paz_adaptive_2004}
Cantu-Paz, E. (2004).
\newblock Adaptive {Sampling} for {Noisy} {Problems}.
\newblock In {\em Genetic and Evolutionary Computation Conference}, volume
  3102, pages 947--958.

\bibitem[Chatzilygeroudis et~al., 2020]{chatzilygeroudis_survey_2020}
Chatzilygeroudis, K., Vassiliades, V., Stulp, F., Calinon, S., and Mouret,
  J.-B. (2020).
\newblock A {Survey} on {Policy} {Search} {Algorithms} for {Learning} {Robot}
  {Controllers} in a {Handful} of {Trials}.
\newblock {\em IEEE Transactions on Robotics}, 36:328--347.

\bibitem[Cully, 2019]{cully2019autonomous}
Cully, A. (2019).
\newblock Autonomous skill discovery with quality-diversity and unsupervised
  descriptors.
\newblock In {\em Genetic and Evolutionary Computation Conference}, pages
  81--89.

\bibitem[Cully et~al., 2015]{cully_robots_2015}
Cully, A., Clune, J., Tarapore, D., and Mouret, J.-B. (2015).
\newblock Robots that can adapt like animals.
\newblock {\em Nature}, 521(7553):503--507.

\bibitem[Cully and Demiris, 2017]{cully_quality_2017}
Cully, A. and Demiris, Y. (2017).
\newblock Quality and {Diversity} {Optimization}: {A} {Unifying} {Modular}
  {Framework}.
\newblock {\em arXiv:1708.09251}.

\bibitem[Cully and Demiris, 2018]{cully_hierarchical_2018}
Cully, A. and Demiris, Y. (2018).
\newblock Hierarchical {Behavioral} {Repertoires} with {Unsupervised}
  {Descriptors}.
\newblock {\em Genetic and Evolutionary Computation Conference}, pages 69--76.

\bibitem[Cully and Mouret, 2015]{cully_evolving_2015}
Cully, A. and Mouret, J.-B. (2015).
\newblock Evolving a {Behavioral} {Repertoire} for a {Walking} {Robot}.
\newblock {\em Evolutionary Computation}, 24(1):59--88.

\bibitem[Duarte et~al., 2018]{duarte2018evolution}
Duarte, M., Gomes, J., Oliveira, S.~M., and Christensen, A.~L. (2018).
\newblock Evolution of repertoire-based control for robots with complex
  locomotor systems.
\newblock {\em IEEE Transactions on Evolutionary Computation}, 22(2):314--328.

\bibitem[Gaier et~al., 2018]{gaier_data-efficient_2018}
Gaier, A., Asteroth, A., and Mouret, J.-B. (2018).
\newblock Data-{Efficient} {Design} {Exploration} through
  {Surrogate}-{Assisted} {Illumination}.
\newblock {\em Evolutionary Computation}, 26(3):381--410.

\bibitem[Gaier et~al., 2020]{gaier2020automating}
Gaier, A., Asteroth, A., and Mouret, J.-B. (2020).
\newblock Automating representation discovery with map-elites.
\newblock {\em Genetic and Evolutionary Computation Conference}.

\bibitem[Gomes et~al., 2018]{gomes_approach_2018}
Gomes, J., Oliveira, S.~M., and Christensen, A.~L. (2018).
\newblock An approach to evolve and exploit repertoires of general robot
  behaviours.
\newblock {\em Swarm and Evolutionary Computation}, 43:265--283.

\bibitem[Gravina et~al., 2019]{gravina_procedural_2019}
Gravina, D., Khalifa, A., Liapis, A., Togelius, J., and Yannakakis, G.~N.
  (2019).
\newblock Procedural {Content} {Generation} through {Quality} {Diversity}.
\newblock In {\em 2019 {IEEE} {Conference} on {Games} ({CoG})}, pages 1--8.

\bibitem[Jin and Branke, 2005]{yaochu_jin_evolutionary_2005}
Jin, Y. and Branke, J. (2005).
\newblock Evolutionary optimization in uncertain environments-a survey.
\newblock {\em IEEE Transactions on Evolutionary Computation}, 9(3):303--317.

\bibitem[Justesen et~al., 2019]{justesen_map-elites_2019}
Justesen, N., Risi, S., and Mouret, J.-B. (2019).
\newblock {MAP}-{Elites} for noisy domains by adaptive sampling.
\newblock In {\em {Genetic} and {Evolutionary} {Computation} {Conference}},
  pages 121--122.

\bibitem[Kruisselbrink et~al., 2010]{kruisselbrink_archive_2010}
Kruisselbrink, J., Emmerich, M., and Bäck, T. (2010).
\newblock An {Archive} {Maintenance} {Scheme} for {Finding} {Robust}
  {Solutions}.
\newblock In {\em Parallel {Problem} {Solving} from {Nature}}, pages 214--223.

\bibitem[Lee et~al., 2018]{Lee2018}
Lee, J., Grey, M.~X., Ha, S., Kunz, T., Jain, S., Ye, Y., Srinivasa, S.~S.,
  Stilman, M., and Liu, C.~K. (2018).
\newblock {DART}: Dynamic animation and robotics toolkit.
\newblock {\em The Journal of Open Source Software}, 3(22):500.

\bibitem[Lehman and Stanley, 2011]{lehman_evolving_2011}
Lehman, J. and Stanley, K. (2011).
\newblock Evolving a diversity of creatures through novelty search and local
  competition.
\newblock In {\em Genetic and Evolutionary Computation Conference}, pages
  211--218.

\bibitem[Miller and Goldberg, 1996]{miller1996genetic}
Miller, B.~L. and Goldberg, D.~E. (1996).
\newblock Genetic algorithms, selection schemes, and the varying effects of
  noise.
\newblock {\em Evolutionary computation}, 4(2):113--131.

\bibitem[Mouret and Clune, 2015]{mouret_illuminating_2015}
Mouret, J.-B. and Clune, J. (2015).
\newblock Illuminating search spaces by mapping elites.
\newblock {\em arXiv:1504.04909}.

\bibitem[Mouret and Doncieux, 2010]{Mouret2010}
Mouret, J.-B. and Doncieux, S. (2010).
\newblock {SFERES}v2: Evolvin' in the multi-core world.
\newblock In {\em Congress on Evolutionary Computation (CEC)}, pages
  4079--4086.

\bibitem[Nguyen et~al., 2016]{nguyen2016understanding}
Nguyen, A., Yosinski, J., and Clune, J. (2016).
\newblock Understanding innovation engines: Automated creativity and improved
  stochastic optimization via deep learning.
\newblock {\em Evolutionary computation}, 24(3):545--572.

\bibitem[Pugh et~al., 2016]{pugh_quality_2016}
Pugh, J.~K., Soros, L.~B., and Stanley, K.~O. (2016).
\newblock Quality {Diversity}: {A} {New} {Frontier} for {Evolutionary}
  {Computation}.
\newblock {\em Frontiers in Robotics and AI}, 3.

\bibitem[Rakshit et~al., 2017]{rakshit_noisy_2017}
Rakshit, P., Konar, A., and Das, S. (2017).
\newblock Noisy evolutionary optimization algorithms – {A} comprehensive
  survey.
\newblock {\em Swarm and Evolutionary Computation}, 33:18--45.

\end{thebibliography}

\end{document}